\documentclass[10pt,twocolumn,letterpaper]{article}

\usepackage{cvpr}              

\usepackage{graphicx}
\usepackage{amsmath}
\usepackage{amssymb}
\usepackage{booktabs}

\usepackage[pagebackref,breaklinks,colorlinks]{hyperref}

\usepackage[capitalize]{cleveref}
\crefname{section}{Sec.}{Secs.}
\Crefname{section}{Section}{Sections}
\Crefname{table}{Table}{Tables}
\crefname{table}{Tab.}{Tabs.}

\def\cvprPaperID{*****} 
\def\confName{CVPR}
\def\confYear{2022}

\begin{document}

\title{6th Place Solution to Google Universal Image Embedding}

\author{Socratis Gkelios\\
Department of Electrical\\ and Computer Engineering\\
Democritus University of Thrace, Greece\\
{\tt\small sgkelios@ee.duth.gr}
\and
Anestis Kastellos and Savvas A. Chatzichristofis\\
Intelligent Systems Laboratory\\ Department of Computer Science\\
Neapolis University Pafos, Cyprus\\
{\tt\small  \{kastellosa\},\{s.chatzichristofis\}@nup.ac.cy}
}
\maketitle

\begin{abstract}
 This paper presents the 6th place solution to the Google Universal Image Embedding competition on Kaggle. Our approach is based on the CLIP architecture; a powerful pre-trained model used to learn visual representations from natural language supervision. We also utilized the SubCenter ArcFace loss with dynamic margins to improve the distinctive power of class separability and embeddings. Finally,  a diverse dataset has been created based on the test’s set categories and the leaderboard’s feedback. By carefully crafting a training scheme to enhance transfer learning, our submission scored 0.685 on the private leaderboard.
\end{abstract}

\section{Introduction}
\label{sec:intro}

Even though image retrieval is a well-studied computer vision task, many challenges still need to be addressed to approach humans' perception of visual similarity \cite{GKELIOS2021114940,DBLP:conf/dcoss/GkeliosBC21}. For the first time, Google launched an image retrieval competition, the Google Universal Image Embedding, that focuses on the techniques' universality and generalization ability. 

In this context, the competition has three main challenges:
\begin{itemize}
\item The training dataset is not provided, so the contenders were responsible for assembling their dataset to train their models.
\item The categories of the test distribution significantly differ, and, considering this is an image retrieval task on the instance-level, the model should be able to handle intra-class similarity on many different categories that do not contain similar characteristics. 
\item The upper limit of embedding's size is 64. 
\end{itemize}
We explored instance recognition/retrieval datasets close to the test's distribution categories to tackle the first challenge. We emphasized powerful pre-trained neural network models trained on large-scale datasets for the second challenge. For the final challenge, we utilized Principal Component Analysis (PCA) so as to reduce the dimensionality of the embedding, retaining as much information as possible. Our final solution is based on the CLIP architecture\cite{radford2021learning} trained on the LAION-2B dataset, a subset of LAION-5B\cite{schuhmann2022laion}. Also, we utilized the ArcFace\cite{deng2020sub} loss function, a popular loss function on Kaggle competitions that maximizes the intra-class separability. Specifically, we adopted the SubCenter ArcFace\cite{deng2019arcface}, which is a variant of the original ArcFace, while also using dynamic margins\cite{ha2020google}


\section{Method}
\label{sec:formatting}
\subsection{Model Architecture}

Our approach employed the CLIP architecture proposed for the image-to-text task. Two public implementations leveraged different datasets for the training procedure: OpenAI's CLIP\cite{radford2021learning} and OpenCLIP\cite{wortsman2022robust}. In the first case, the model is pre-trained on Imagenet22K, and the best-performing model is ViT-L, while in the second case, the training was performed on the LAION-2B dataset, with the best-performing model being ViT-H. We used only the image encoder part of the original topology in both cases. The results strongly suggested that the OpenCLIP model was significantly better than openai's, mainly because ViT-H has more learning capacity than ViT-L. So we proceeded with ViT-H CLIP. We also modified the head by taking the output of the projection layer and feeding it to a BN-Dropout-FC block. The dropout rate is set to 0.2, and the FC downsizes the 1024-dimensional embedding to 256. The 256 vector passes through the ArcFace layer to obtain the classification logits.

To handle the instance retrieval task, we used the SubCenter ArcFace. This variant of the original ArcFace improves robustness when noisy training data. This is achieved by relaxing the constraint that every class has a single center by introducing more sub-centers per class. Models trained in this manner can showcase better representation capability and produce high-quality embeddings. Although our training dataset does not contain extreme imbalance, some classes are underrepresented ($\approx$5 images per class) compared to others ($>$ 40 images per class). So dynamic margins are used to alleviate this problem. In essence, the margins are allocated dynamically per class depending on the number of instances representing them. We facilitate the training procedure by assigning bigger margins on underrepresented classes to make separating them easier. We defined the upper bound to 0.45 and the lower bound to 0.005.

\subsection{Datasets}

We explored a plethora of datasets closely related to the test’s distribution. Specifically, we used subsets of the following datasets for our best score: Google LandmarksV2\cite{weyand2020google}, Products10k\cite{bai2020products}, Food-101\cite{bossard14}, iMaterialist, Fashion200k\cite{han2017automatic}, DeepFashion\cite{liu2016deepfashion}, RP2K\cite{peng2020rp2k}, Stanford Cars\cite{KrauseStarkDengFei-Fei_3DRR2013}, Stanford Online Products\cite{song2016deep}, MET Artwork dataset\cite{ypsilantis2021met} and Storefronts-146\footnote{https://www.kaggle.com/datasets/kerrit/storefront-146}. Although iMaterialist does not contain instance labels, we created manually, with the help of the pre-trained CLIP, about 400 additional labeled furniture images. We tried to approximate the percentage of each category on the test set for the most important categories. In our initial experiments, we had around 200k images in our training set but noticed that the dataset size and the public leaderboard score followed a similar trend. As a result, we assembled a dataset of $\approx$655k images from the categories mentioned above. In most datasets, we selected classes randomly with at least three samples.

\section{Experiments}
\subsection{Training details}

The OpenCLIP presents state-of-the-art performance on zero-shot tasks; therefore, the model offers high-quality image descriptors out-of-the-box. The challenge was to train this model without altering its learned weights and causing a significant distribution shift. We froze the CLIP backbone during our first training attempts and trained the head for five epochs with a 0.0001 learning rate (lr). After this step, we unfroze the ViT-H until resblock 15 and then trained for another epoch with the lr reduced by a factor of 10. As we started adding more data to our pipeline, we noticed, we could train the backbone even more, and for this reason, we adopted the following scheme that applies different lr to the backbone and the head,
$$
Learning Rate=
\begin{cases}
1e^{-7},\ for\ model.backbone\\
1e^{-4},\ for\ model.head\\
\end{cases}
$$

and trained the model for four epochs. Again we only unfroze the model from resblock31 to resblock 15 for three main reasons:

\begin{itemize}
\item We did not observe improvement by unfreezing more blocks
\item Computational complexity and training time
\item Initial layers usually contain low-level generic features, so we did not want to alter the learned features from a bigger dataset
\end{itemize}
Remarkably, this scheme performed worse when the smaller dataset was used. Therefore, we set the batch size to 32 and employed the Adam optimizer.  

We used the following augmentations during training inspired by past Kaggle instance level competitions:
Horizontal flip, image compression, shift, scale, rotate, cutout, random brightness, contrast, and RGB-shift.
\subsection{Validation strategy}

We attempted to create a validation set with classes that were not seen in training, with and without matching the distribution of the test set. After extracting the image descriptors (with pooling), we trained a NearestNeighbor algorithm that generated a ranking list of the closest matches. We evaluated the ranking list using the mAP metric. Unfortunately, none of our validation schemes correlated adequately with the public leaderboard, so technically, our only validation set was the public leaderboard. We used the local validation only to estimate if there was a problem with the new approaches. Also, we measure the per-vertical mAP for each category to see if we can infer any additional insights on which category our model seems to struggle with.

\subsection{Inference}

During inference, the images were resized to 224$\times$224 to match the model's expected input. We also normalized the images according to our training pipeline. Finally, we tested two approaches for reducing the dimensionality of the final vector to 64 as the model output a 256-dimensional vector. The first was simply using average pooling, while in the second, the features were transformed using PCA. We fitted PCA  on a public dataset shared on Kaggle\footnote{https://www.kaggle.com/datasets/rhtsingh/130k-images-512x512-universal-image-embeddings} comprised of 130k images from all categories of the test set. The performance with PCA was significantly better than the average pooling by 0.016 on the public leaderboard and 0.005 on the private set. The final solution was comprised of a single model without


\section{Conclusions}
This paper presented our approach to the Google Universal Embedding Challenge. We showed that by incorporating different lr to the backbone and the head, the performance of the pre-trained CLIP model could be improved significantly. Additionally, PCA can be utilized effectively to reduce the dimensionality of the final vector by providing a robust embedding.

{\small
\bibliographystyle{ieee_fullname}
\bibliography{egbib}
}

\end{document}